\documentclass[10pt,twocolumn,letterpaper]{article}

\usepackage[pagenumbers]{cvpr} % To force page numbers, e.g. for an arXiv version

\usepackage{amsmath,amsfonts,bm}
\usepackage[table,xcdraw,dvipsnames]{xcolor}
\usepackage{colortbl}
\usepackage{pifont}
\usepackage{multirow}

\usepackage{url}

\usepackage{graphicx}
\usepackage{amsmath}
\usepackage{amssymb}
\usepackage{mathrsfs}
\usepackage{booktabs}
\usepackage{comment}
\usepackage{color}

\usepackage{wrapfig}
\usepackage{algorithm}
\usepackage{algpseudocode}

\usepackage{latexsym}
\usepackage{makecell,diagbox}
\usepackage{algorithm}
\usepackage{listings}

\usepackage{amsfonts}
\usepackage{colortbl}
\usepackage{microtype}
\usepackage{multirow}
\usepackage{adjustbox}
\usepackage{enumitem}
\usepackage{float}
\usepackage{bm}
\usepackage{bbm} % \mathbbm
\usepackage{etoc}
\usepackage{bbding}
\usepackage{pifont}
\usepackage{wasysym}
\usepackage{xcolor}

\newcommand{\cmark}{\textcolor{OliveGreen}{\ding{51}}}
\newcommand{\xmark}{\textcolor{BrickRed}{\ding{55}}}
\def\eqref#1{equation~\ref{#1}}
\def\1{\bm{1}}

\DeclareMathAlphabet{\mathsfit}{\encodingdefault}{\sfdefault}{m}{sl}
\SetMathAlphabet{\mathsfit}{bold}{\encodingdefault}{\sfdefault}{bx}{n}

\definecolor{cvprblue}{rgb}{0.21,0.49,0.74}
\usepackage[pagebackref,breaklinks,colorlinks,citecolor=cvprblue]{hyperref}

\def\paperID{10365} % *** Enter the Paper ID here
\def\confName{CVPR}
\def\confYear{2024}

\title{HAtt-Flow: Hierarchical Attention-Flow Mechanism for \\ Group Activity Scene Graph Generation in Videos}

\author{Naga VS Raviteja Chappa$^{1}$, Pha Nguyen$^{1}$, Thi Hoang Ngan Le$^{1}$,Khoa Luu$^{1}$\\
$^{1}$University of Arkansas\\
\tt\small \{nchappa, panguyen, thile, khoaluu\}@uark.edu,  
\vspace{-2mm}
}

\begin{document}
\twocolumn[{
\renewcommand\twocolumn[1][]{#1}%
\maketitle
\begin{center}
 \centering
 \vspace{-2mm}
 \captionsetup{type=figure}
\includegraphics[width=\textwidth]{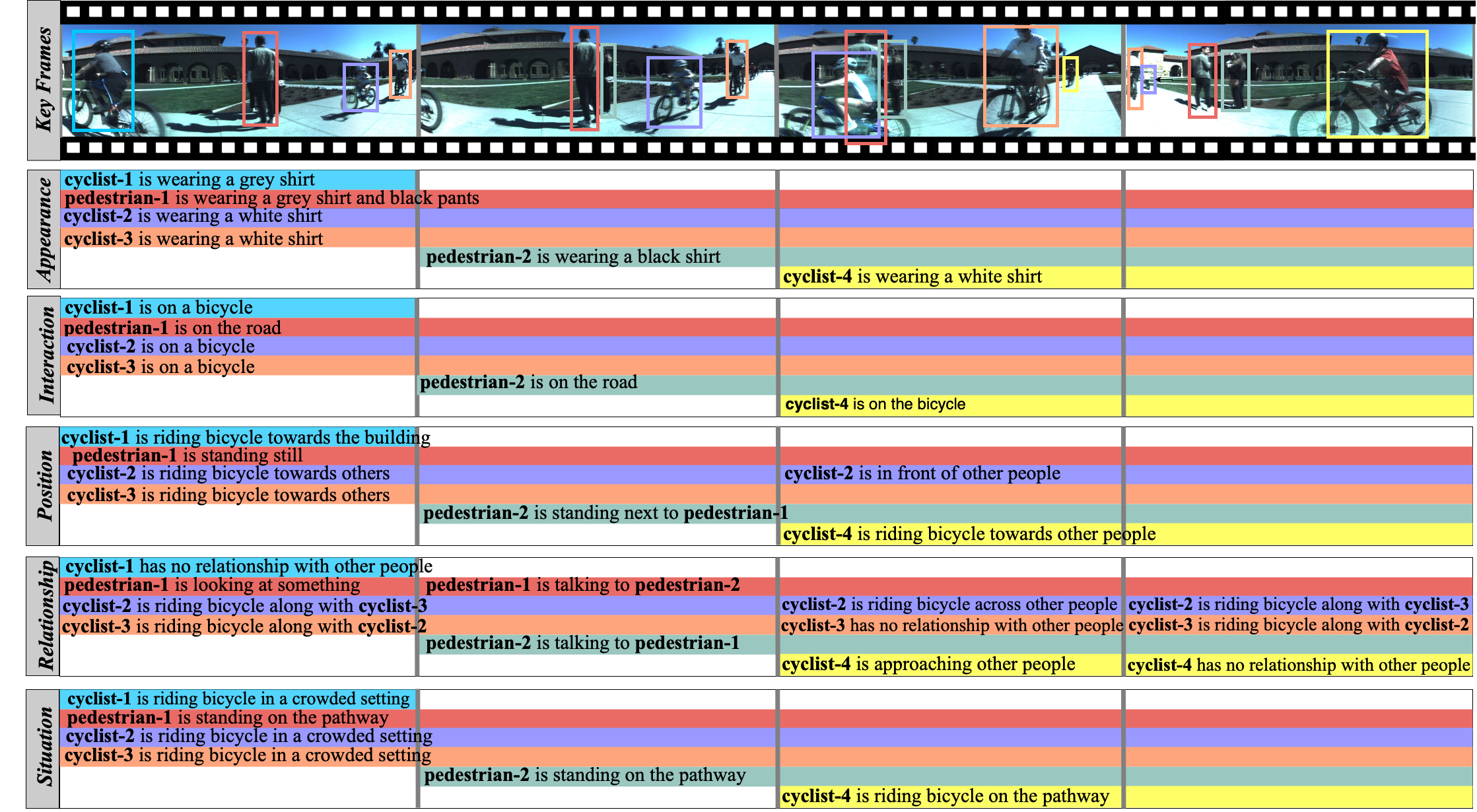}
 % \vspace{1mm}
 \captionof{figure}{\textbf{A sample video from our Group Activity Scene Graph (GASG) Dataset.} The top row displays keyframes featuring overlaid bounding boxes, each annotated with a unique ID for consistency. Below, the timeline tubes provide a comprehensive temporal representation of scene graph annotations for distinct attributes, including \textit{\textbf{Appearance, Interaction, Position, Relationship, and Situation}}. These annotations offer nuanced details, enhancing scene understanding and contributing to a more refined video content analysis. \textbf{Best viewed in color and zoomed.}}
 \label{fig:page1}
\end{center}
}]

\begin{abstract}

%\textcolor{red}{[@KL (11/3/23: , good progress!.]}

%\textcolor{red}{[@KL (10/28/23: Ravi, Can you complete the Introduction and Literature Review Sections this weekend?]}

%\textcolor{red}{[@KL (10/15/23: Ravi, let's spend time on drafting the paper.]}

%\textcolor{red}{[@KL: Ravi, pls focus on the section of the proposed method first.]}

%\textcolor{red}{[@KL: Let's try to complete the first draft by today, Friday. I will start revising the paper this weekend.] }

%\textcolor{red}{[@KL: Consider using better viewable color tone in the figures.] }

Group Activity Scene Graph (GASG) generation is a challenging task in computer vision, aiming to anticipate and describe relationships between subjects and objects in video sequences. Traditional Video Scene Graph Generation (VidSGG) methods focus on retrospective analysis, limiting their predictive capabilities. To enrich the scene understanding capabilities, we introduced a GASG dataset extending the JRDB dataset with nuanced annotations involving \textit{Appearance, Interaction, Position, Relationship, and Situation} attributes. This work also introduces an innovative approach, \textbf{H}ierarchical \textbf{Att}ention-\textbf{Flow} (HAtt-Flow) Mechanism, rooted in flow network theory to enhance GASG performance. Flow-Attention incorporates flow conservation principles, fostering competition for sources and allocation for sinks, effectively preventing the generation of trivial attention.
Our proposed approach offers a unique perspective on attention mechanisms, where conventional "values" and "keys" are transformed into sources and sinks, respectively, creating a novel framework for attention-based models.
% To address the specific challenges of the GASG task, we introduce customized evaluation metrics that provide a nuanced assessment of model performance.
Through extensive experiments, we demonstrate the effectiveness of our Hatt-Flow model and the superiority of our proposed Flow-Attention mechanism. This work represents a significant advancement in predictive video scene understanding, providing valuable insights and techniques for applications that require real-time relationship prediction in video data.

\end{abstract}

\section{Introduction}\label{sec:intro}

Visual scene understanding is a foundational challenge in computer vision, encompassing the interpretation of complex scenes, objects, and their relationships within images and videos. This task is particularly intricate in video, where temporal dynamics and multi-modal information introduce unique complexities. Group Activity Video Scene Graph (GAVSG) generation, which involves predicting relationships between objects in a video across multiple frames, stands at the forefront of this endeavor.

In recent years, significant progress has been made in video understanding. Techniques such as Video Scene Graph Generation (VidSGG) have allowed us to extract high-level semantic representations from video content. However, VidSGG typically operates in a static, retrospective manner, constraining its predictive capabilities. The GAVSG dataset, on the other hand, extends the scope of visual scene understanding to anticipate and describe subject and object relationships and their temporal evolution.

In the closely-linked domain of human-object interaction (HOI)\cite{gupta2015visual}, transferable techniques have proven effective for scene graph generation (SGG) tasks~\cite{gkioxari2018detecting, kato2018compositional, chao2018learning, wang2019deep, li2019transferable, zhou2020cascaded, wang2020learning, hou2020visual, li2020detailed, gao2020drg, kim2020uniondet, liu2020amplifying, tamura2021qpic, hou2021affordance, zhang2021mining, wang2022learning, zhang2022efficient}, some of which inspired the foundation of PSG dataset baselines~\cite{kim2021hotr, zou2021end}.

\begin{figure}[t!]
    \centering
    \includegraphics[width=0.47\textwidth]{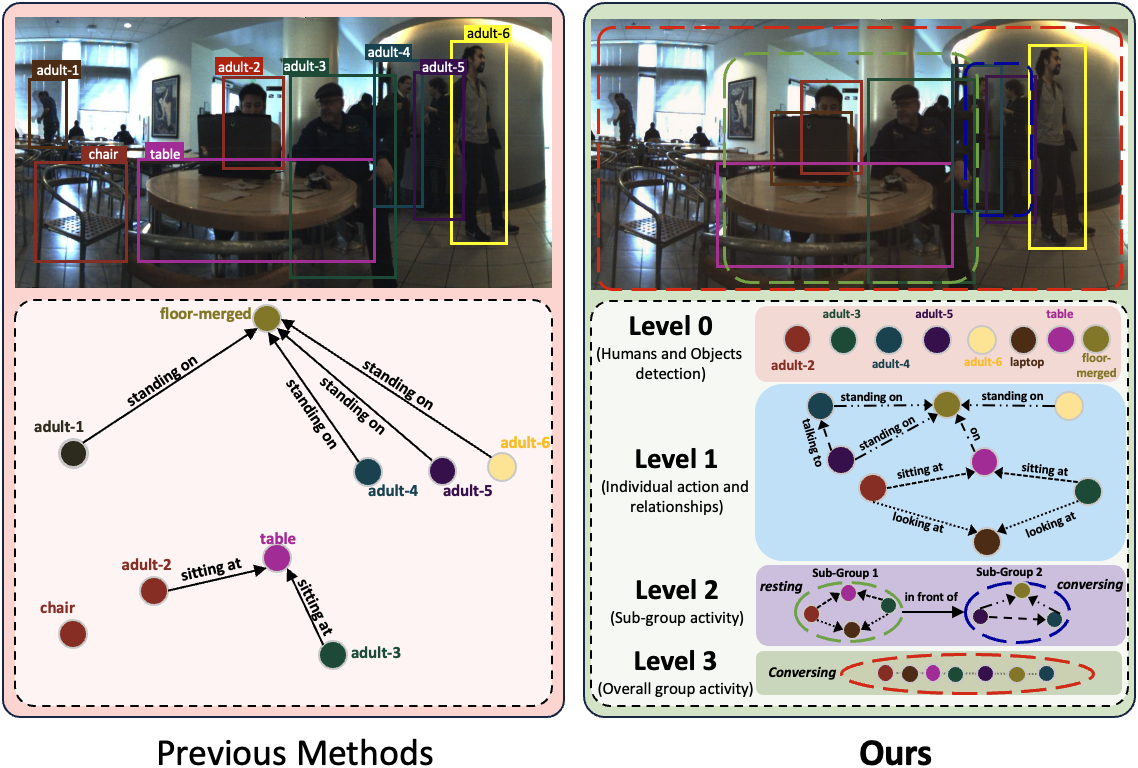}
    \caption{
    {{Comparison of HAtt-Flow result with other Scene Graph Generation methods.} \textbf{Best viewed in color and zoomed.}}}
    \label{fig:motivation}
\end{figure}

In the context of SGG, diverse methodologies have been explored, from probabilistic graphical models and AND-OR grammar approaches~\cite{amer2012cost, amer2013monte, amer2014hirf, amer2015sum, lan2011discriminative, lan2012social, shu2015joint, wang2013bilinear} to knowledge graph embeddings like VTransE~\cite{zhang2017visual} and UVTransE~\cite{hung2020contextual}. Recent endeavors delve into challenges such as the long-tailed distribution of predicates~\cite{tang2020unbiased, desai2021learning}, visually-irrelevant predicates~\cite{liang2019vrr}, and precise bounding box localization~\cite{khandelwal2021segmentation}. As shown in the~\cref{fig:motivation}, we can observe that the previous methods can detect the subjects and objects in the scene. However, they need to generate a well-defined scene graph, whereas our method can learn all the nuanced relationships among the subjects and objects in the scene to produce a fine scene graph.

To address the limitation of enriched-learning of the group activities in the scene, we introduced the GASG dataset which includes nuanced annotations in the form of five different attributes. This can help to set a better scene graph generation benchmark than the existing datasets in this domain.
In this work, we propose a novel approach for GAVSG that draws inspiration from flow network theory, introducing Flow-Attention. This mechanism leverages flow conservation principles in both source and sink aspects, introducing a competitive mechanism for sources and an allocation mechanism for sinks. This innovative approach mitigates the generation of trivial attention and enhances the predictive power of GAVSG.
We build upon a new perspective on attention mechanisms, rooted in flow network theory, to design our GAVSG framework. The conventional attention mechanism aggregates information from "values" and "keys" based on the similarity between "queries." By framing attention in terms of flow networks, we transform values into sources and keys into endpoints, thus creating a fresh perspective on the attention mechanism.

% Our contributions extend to the design of novel evaluation metrics tailored to the GAVSG task, providing a more nuanced assessment of model performance. These metrics address the challenges of handling repetitive or scattered relationship descriptions, a departure from conventional metrics.

\noindent\textbf{Contributions:} The main contributions of our work are three folds. First, we introduce a novel dataset\footnote{The dataset is available for verification in the supplementary material.} with nuanced attributes that aid the scene graph generation task in the group activity setting. Second, our work advances the state of the art in predictive video scene understanding by introducing Flow-Attention and redefining attention mechanisms by incorporating hierarchy awareness. Third, we demonstrate the effectiveness of our approach through extensive experiments and achieve State-of-the-Art performance over the existing approaches.

\begin{table}[t!]
\caption{Comparison of existing datasets. \textbf{GA} is the group activity label, \textbf{H-H}, \textbf{H-O} and \textbf{O-O} represents interaction between \emph{Human and Human, Human and Object \& Object and Object}.}
 \resizebox{\columnwidth}{!}{
\begin{tabular}{c|c|ccc|ccc}
\hline
\multirow{2}{*}{\textbf{Datasets}}  & \multirow{2}{*}{\textbf{Settings}} & \multicolumn{3}{c|}{\textbf{Annotations}}                             & \multicolumn{3}{c}{\textbf{Attributes}}                               \\ \cline{3-8} 
                                   &                                                                          & \textbf{BBox}         & \textbf{IDs}          & \textbf{GA}           & \textbf{H-H}          & \textbf{H-O}          & \textbf{O-O}          \\ \hline
ActivityNet~\cite{caba2015activitynet}                                                       & 1                                  & \cmark & \xmark & \xmark & \xmark & \cmark & \xmark \\
MSRVTT~\cite{xu2016msr}                                                                & 1                                  & \cmark & \xmark & \xmark & \xmark & \cmark & \xmark \\
MSVD~\cite{pan2017video}                                                               & 1                                  & \cmark & \xmark & \xmark & \xmark & \cmark & \cmark \\
PSG~\cite{yang2022panoptic}                                                             & 1                                  & \cmark & \cmark & \xmark & \cmark & \cmark & \cmark \\
PVSG~\cite{yang2023panoptic}                                                             & 1                                  & \cmark & \cmark & \xmark & \cmark & \cmark & \cmark \\ \hline
\textbf{GASG (Ours)}                                               & \textbf{5}                                  & \cmark & \cmark & \cmark & \cmark & \cmark & \cmark \\ \hline
\end{tabular}} \label{tab:comparison}
\end{table}

\section{Related Work}\label{sec:related}
% \textcolor{blue}{Table comparison} 
% \textcolor{red}{- why flow attention is better than text-video alignment?
% -have comparisons dicussion between existing VLMs and SOTAs}

% \noindent\textcolor{red}{Section about GAR} 
\noindent\textbf{Group Action Recognition (GAR).}
Group Action Recognition (GAR) has witnessed a shift towards deep learning methodologies, notably convolutional neural networks (CNN) and recurrent neural networks (RNN)\cite{bagautdinov2017social, deng2016structure, ibrahim2016hierarchical, ibrahim2018hierarchical, li2017sbgar, qi2018stagnet, shu2019hierarchical, wang2017recurrent, yan2018participation}. Attention-based models and graph convolution networks are crucial in capturing spatial-temporal relations in group activities. Transformer-based encoders, often coupled with diverse backbone networks, excel in extracting features for discerning actor interactions in multimodal data\cite{gavrilyuk2020actor}. Recent innovations, such as MAC-Loss, introduce dual spatial and temporal transformers for enhanced actor interaction learning~\cite{han2022dual}. The field continues to evolve with heuristic-free approaches like those by Tamura et al., simplifying the process of social group activity recognition and member identification~\cite{tamura2022hunting}.

\noindent\textbf{Scene Graph Generation (SGG).}
In Scene Graph Generation (SGG), the traditional two-stage paradigm involves object detection and pairwise predicate estimation~\cite{johnson2015image, dai2017detecting, zhang2017relationship, kolesnikov2019detecting, qi2019attentive, xu2017scene, zellers2018neural, tang2018vctree, yang2018graph, lin2020gps, chen2019knowledge, li2018factorizable}. Recent advancements include knowledge graph embeddings, graph-based architectures, energy-based models, and linguistic supervision~\cite{suhail2021energybased, tang2018vctree, gu2019scene, zareian2020bridging, ZareianWYC20, lu2016visual, zhong2021learning, ye2021linguistic}. To address challenges like long-tailed distribution and visually irrelevant predicates, the field has seen a pivot towards panoptic segmentation-based SGG, inspired by the simultaneous generation of scene graphs and semantic segmentation masks~\cite{khandelwal2021segmentation}.
Notably, insights from the closely-linked domain of human-object interaction (HOI) have influenced SGG techniques~\cite{gkioxari2018detecting, kato2018compositional, chao2018learning, wang2019deep, li2019transferable, zhou2020cascaded, wang2020learning, hou2020visual, li2020detailed, gao2020drg, kim2020uniondet, liu2020amplifying, tamura2021qpic, hou2021affordance, zhang2021mining, wang2022learning, zhang2022efficient}.

\noindent\textbf{Video Scene Graph Generation (VidSGG).}
VidSGG, initiated by Shang \etal~\cite{shang2017video}, explores spatio-temporal relations in videos. Research has delved into spatio-temporal conditional bias, domain shift between image and video scene graphs, and embodied semantic approaches using intelligent agents. Notable methods include TRACE~\cite{teng2021target}, which separates relation prediction and context modeling, and Embodied Semantic SGG, employing reinforcement learning for path generation by intelligent agents~\cite{dong2019baconian,li2022embodied}. Yang \etal~\cite{yang2023panoptic} proposed a transformer encoder based baseline model to evaluate their proposed panoptic video scene graph dataset which includes fusing the extracted features of the subjects in the scene.

\subsection{Limitation of Prior Datasets}
% \textcolor{red}{[@KL (10/28/23: Need updating.]}
We presented a detailed comparison of existing datasets in ~\cref{tab:comparison}. The limitations of prior datasets become particularly conspicuous when we contemplate the intricate nature of group activity within the visual content. Most of these datasets have predominantly concentrated on specific individual action types, often neglecting the complex nature of group activities in real-world interactions. This constrained focus has created obstacles in developing models capable of addressing a wide range of action classifications, limiting their adaptability to diverse real-world scenarios. Furthermore, earlier datasets often underscore relationships within isolated fragments of the relational graph, occasionally omitting the complexities of more extensive and intricate scenes. Some of these datasets have suffered from sparse annotations, which, at times, have led to a lack of comprehensive relationship modeling and potential model bias. 

Notably, there remains to be a significant void in the representation of scenes featuring dense crowds of people. These settings bring about formidable challenges related to occlusion management and a nuanced understanding of complex interactions within such contexts. 

In response to these limitations, we introduce \textit{Group Activity Scene Graph} (GASG) dataset that directly addresses these issues. The dataset boasts various scenes and settings (which are represented as attributes in the annotations), spanning distinct scenarios, effectively differentiating it from the prior datasets. It excels in capturing five critical features in group activity: appearance, situation, position, interaction, and relations. Furthermore, this dataset comprehensively tracks the movements and interactions of individuals and sub-groups, allowing for a profound understanding of their dynamics and activities over time. With a rich dataset on these aspects, this {GASG} dataset lays the groundwork for a new paradigm in understanding complex group activities within scenarios characterized by dense populations, thereby pushing the boundaries of group activity recognition.

\section{Dataset Overview}
\subsection{Data Collection and Annotation}\label{sec:dataset_collection}

This {GASG} dataset comprises a rich collection of videos from the JRDB dataset, offering a unique perspective on sub-group and overall group activities. This dataset provides comprehensive coverage of various activities and introduces essential tracking information.

The tracking information within the {GASG} dataset facilitates a detailed understanding of individuals and sub-groups within each frame. This information includes the trajectory, position, and interactions of each actor. The dataset defines five key aspects for comprehensive scene understanding:
\begin{itemize}
    
\item \textit{Interaction}: This aspect characterizes the dynamic interactions between subjects and objects, shedding light on how individuals and sub-groups engage with each other.

\item \textit{Position}: The dataset includes precise data on the location and orientation of subjects and objects, enhancing the analysis of their spatial relationships during activities.

\item \textit{Appearance}: Visual traits of subjects and objects are meticulously captured, allowing for detailed examinations of their attributes and characteristics.

\item \textit{Relationship}: Understanding the associations and connections between subjects and objects is essential for deciphering the complex interplay within group activities. This aspect provides insight into the underlying dynamics of relationships within the scenes.

\item \textit{Situation}: To provide environmental context, the {GASG} dataset offers descriptors highlighting the contextual information surrounding subjects and objects, enabling researchers to consider the broader setting in their analyses.

\end{itemize}
\begin{figure}[t!]
    \centering
    \includegraphics[width=0.49\textwidth]{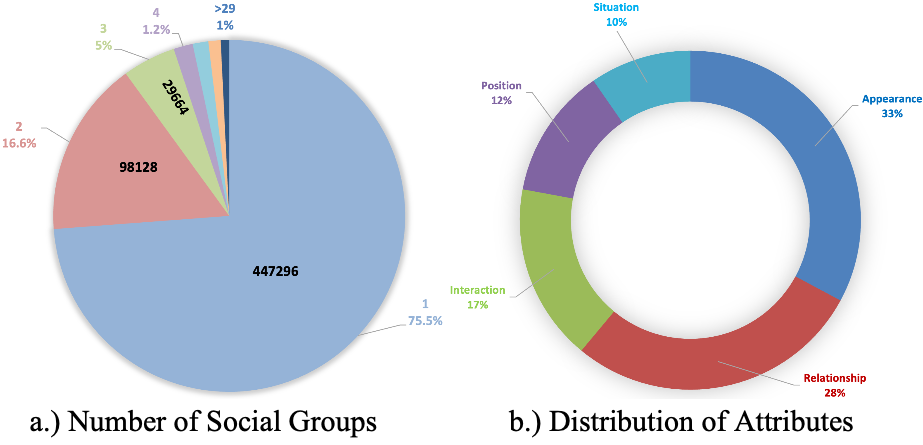}
    \caption{
    {{Statistics of the GASG dataset, number of social groups, and the attributes in the dataset.} \textbf{Best viewed in color and zoomed.}}}
    \label{fig:stats}
\end{figure}
These five key aspects—Interaction, Position, Appearance, Relationship, and Situation—form the backbone of the dataset's annotation structure, providing a holistic view of the diverse activities and interactions within the sub-group and overall group scenarios. This level of detail sets {GASG} apart, making it a valuable resource for research in scene comprehension, action recognition, and group activity analysis. The annotation process is detailed in the \textit{Appendix}.

\subsection{Dataset Statistics}
The accompanying pie chart in~\cref{fig:stats} b.) delves into the complexity of attributes in our dataset, revealing that "Appearance" consists of 33\% of the total annotations, which is the highest, "Relationship" consists of 28\%, "Interaction" consists of 17\%, "Position" consists of 12\% and "Situation" consists of 10\% which is the least.
We explore the distribution of social activity labels in ~\cref{fig:stats} a.), focusing on the sizes of social groups. The chart in the figure provides a nuanced view of social group sizes in the dataset. Specifically, 75.5\%, 16.6\%, 5\%, and 1.2\% of social groups consist of one, two, three, and four members, respectively. Interestingly, only 1\% of the dataset includes groups with five or more members, with the maximum observed group size being 29 members.
% \subsection{Benchmarking Protocols}

% \textcolor{red}{[@KL (11/4/23: Need updating.]}

% \subsection{Metrics}

% \textcolor{red}{[@KL (11/4/23: Need updating.]}
\begin{figure*}[t!]
    \centering
    \includegraphics[width=1.0\textwidth]{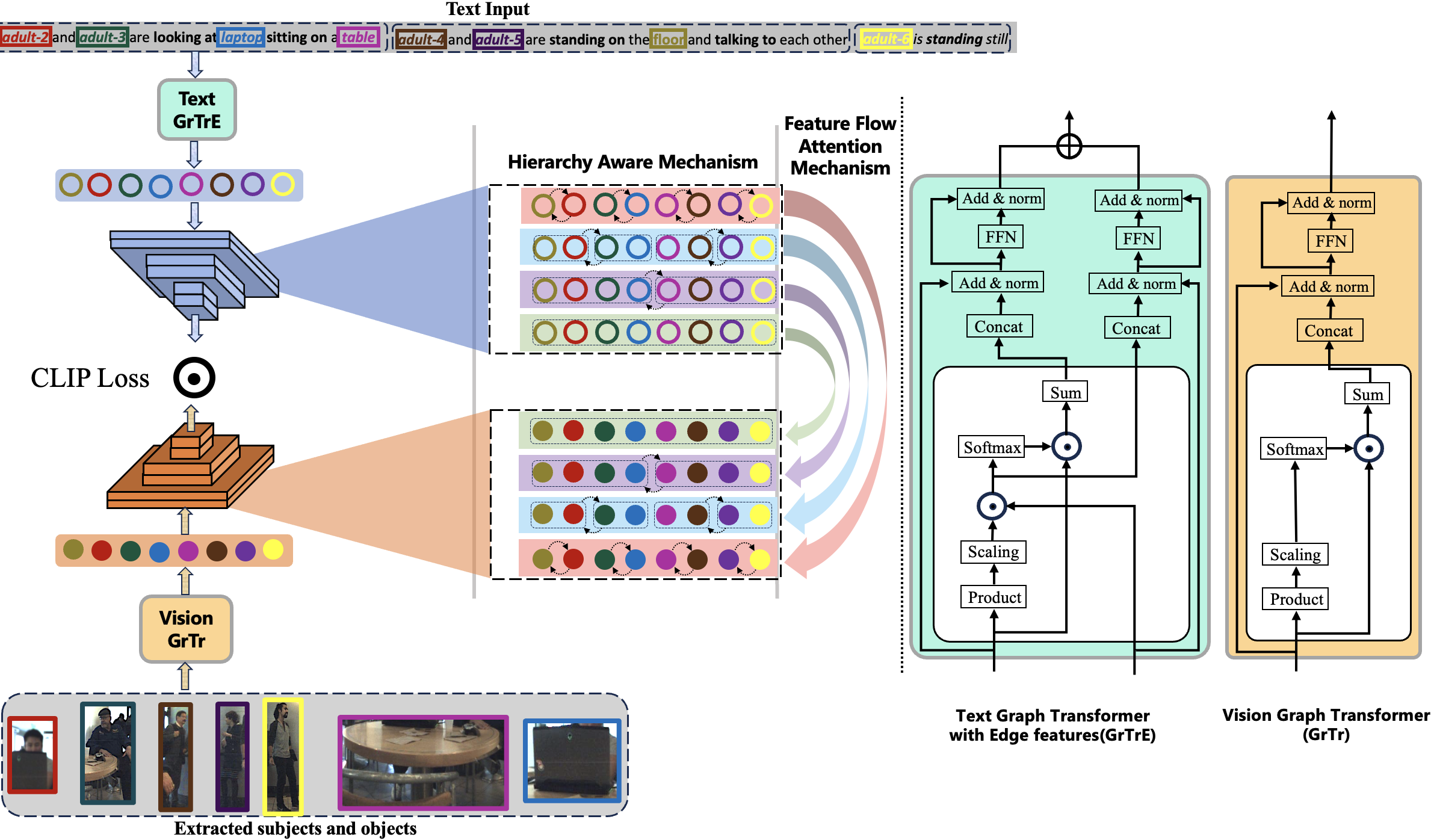}
     \put(-468,146){\normalsize($\mathscr{L}$)}
     \put(-417,145){$\widehat{\mathbf{v}^{'}}$}
     \put(-417,168){$\widehat{\mathbf{t}^{'}}$}
     \put(-312,286){(\small$\mathbf{t}$)}
     \put(-349,1){($\mathbf{v}$)}
     \put(-385,225){$\mathbf{t^{'}}$}
     \put(-389,95){$\mathbf{v^{'}}$}
     \put(-50,53){\small$\mathbf{v}$}
     \put(-37,255){\small$\mathbf{v^{'}}$}
     \put(-127,255){\small$\mathbf{t^{'}}$}
     \put(-152,52){\small$\mathbf{t}$}
     \put(-102,52){\small$\mathbf{t}_e$}
     \put(-325,220){\tiny$\mathbf{L}_0$}
     \put(-325,205){\tiny$\mathbf{L}_1$}
     \put(-325,190){\tiny$\mathbf{L}_2$}
     \put(-325,175){\tiny$\mathbf{L}_3$}
     \put(-325,140){\tiny$\mathbf{L}_3$}
     \put(-325,125){\tiny$\mathbf{L}_2$}
     \put(-325,110){\tiny$\mathbf{L}_1$}
     \put(-325,95){\tiny$\mathbf{L}_0$}
     
    \caption{
    \textbf{Overall architecture of the proposed HAtt-Flow network.} The extracted visual and textual features are passed through their respective graph transformers to obtain corresponding node features. These nodes are passed through the hierarchy-aware-based transformer encoder models to have enriched features, including a feature flow attention mechanism to enhance cross-modality learning. Finally, we use CLIP loss to optimize the learned features. Please refer to ~\cref{fig:motivation} for the details of levels $\mathbf{L}_{0}$, $\mathbf{L}_{1}$, $\mathbf{L}_{2}$ and $\mathbf{L}_{3}$.}
    \label{fig:overall}
\end{figure*}
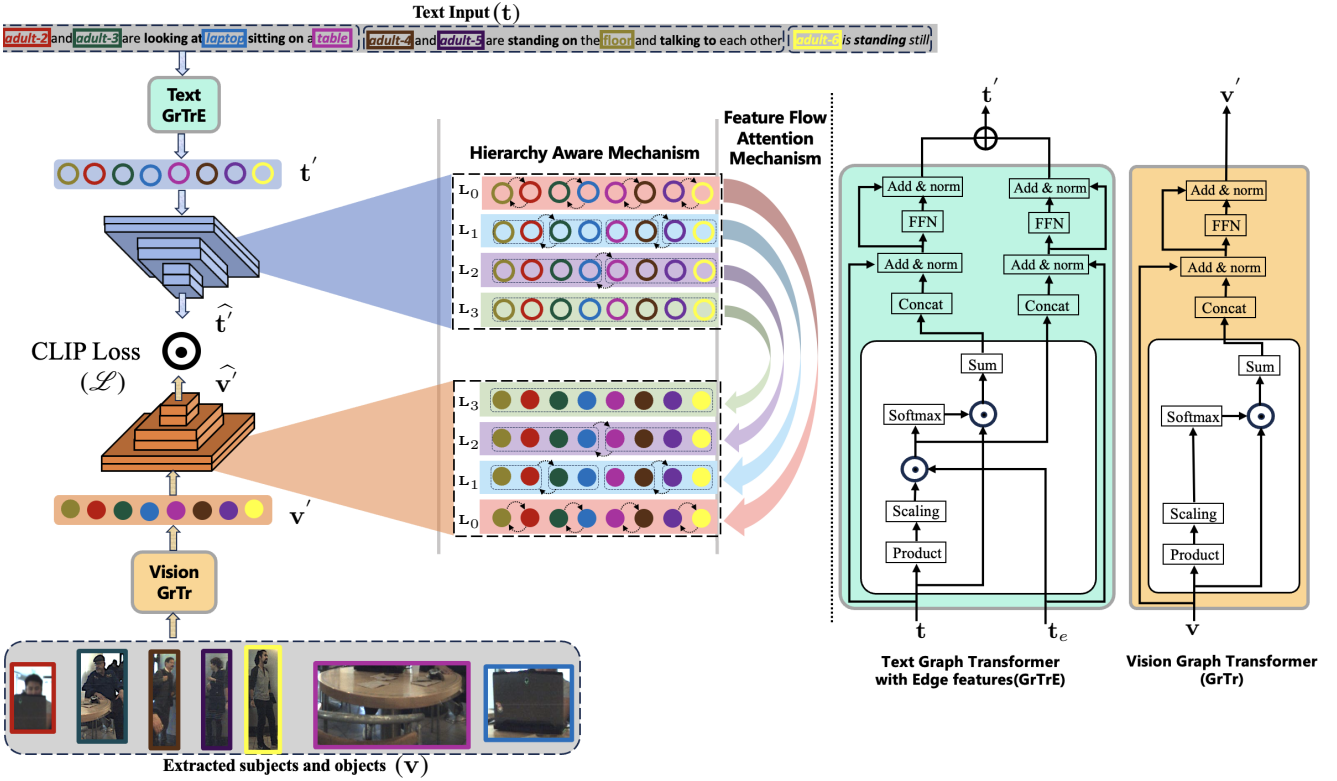
\section{Methodology}\label{sec:methods}

% \textcolor{red}{@KL: Ravi, focus on the proposed method first. Ensure we include the following items: Novelty, theory, contributions and the details of the proposed method.}

We utilized pre-trained visual and textual backbones to extract the corresponding subject features in the video $\mathbf{v}$ and textual features $\mathbf{t}$. In the input preparation section, these are ${h}$ to represent the nodes and ${e}$ to represent the edges. However, we denote the visual nodes as $\mathbf{v}$, textual nodes as $\mathbf{t}$ and textual edges as $\mathbf{t}_{e}$ in~\cref{fig:overall}. We used the Graph Transformer Layer and the Graph Transformer Layer with edge features to extract the corresponding feature representations. The former is tailored for graphs lacking explicit edge attributes, while the latter incorporates a dedicated edge feature pipeline to integrate available edge information, maintaining abstract representations at each layer.

\noindent\textbf{Input Preparation}
Initially, we prepare input node and edge embeddings for the Graph Transformer Layer. In the context of our model, text features are employed to generate both nodes and edges, whereas vision features are exclusively utilized for generating nodes. Consider a graph $\mathcal{G}$ with node features represented as text features $\alpha_i \in \mathbb{R}^{d_n \times 1}$ for each node $i$ and edge features, also derived from text, denoted as $\beta_{ij} \in \mathbb{R}^{d_e \times 1}$ for edges between nodes $i$ and $j$. The input node features $\alpha_i$ and edge features $\beta_{ij}$ undergo a linear projection to be embedded into $d$-dimensional hidden features $h_i^{0}$ and $e_{ij}^{0}$.
\begin{equation}
\label{eqn:input_embd}
\hat{h}i^{0} = A^{0} \alpha_i + a^{0} ; e{ij}^{0} = B^{0} \beta_{ij} + b^{0} ,
\end{equation}
Here, $A^{0} \in \mathbb{R}^{d \times d_n}$, $B^{0} \in \mathbb{R}^{d \times d_e}$, and $a^{0}, b^{0} \in \mathbb{R}^{d}$ are parameters of the linear projection layers. The pre-computed node positional encodings of dimension $k$ are linearly projected and added to the node features $\hat{h}_i^{0}$.
\begin{equation}
\label{eqn:pe_embd_add}
{\lambda}_i^{0} = C^{0} \lambda_i + c^{0} ; h_i^{0} = \hat{h}_i^{0} + {\lambda}_i^{0},
\end{equation}
Here, $C^{0} \in \mathbb{R}^{d \times k}$ and $c^{0} \in \mathbb{R}^{d}$. Notably, positional encodings are only added to the node features at the input layer and not during intermediate graph transformer layers. The detailed information about the Graph transformer layers will be presented in \textit{Appendix}.
\subsection{Hierarchical Awareness Induction}
We propose enriching the vision and language branches through a hierarchy-aware attention mechanism. In line with the conventional transformer architecture, we divide modality inputs into low-level video patches and text tokens. These are recursively merged based on semantic and spatial similarities, gradually forming more semantically concentrated clusters, such as video objects and text phrases. We define hierarchy aggregation priors with the following aspects:

\noindent\textbf{Tendency to merge.} Patches and tokens are recursively merged into higher-level clusters that are \emph{spatially and semantically} similar. If two nearby video patches share similar appearances, merging them is a natural step to convey the same semantic information.

\noindent\textbf{Non-splittable.} Once patches or tokens are merged, they will \emph{never be split} in later layers. This constraint ensures that hierarchical information aggregation never degrades, preserving the complete process of hierarchy evolution layer by layer.

We incorporate these hierarchy aggregation priors into an attention mask $C$, serving as an extra inductive bias to help the conventional attention mechanism in Transformers better explore hierarchical structures adapted to each modality format—2D grid for videos and 1D sequence for texts. Thus, the proposed hierarchy-aware attention is defined as:
\begin{equation}
\emph{{Hierarchy\_Attention}} = \left( C \odot \operatorname{softmax}\left(\frac{Q K^T}{\sqrt{d_h}}\right) \right) V
\end{equation}
Note that $C$ is shared among all heads and progressively updated bottom-up across Transformer layers. We elaborate on the formulations of the hierarchy-aware mask $C$ for each modality as follows.

\noindent\textbf{Hierarchy Induction for Language Branch}

In this section, we reconsider the tree-transformer method from the perspective of the proposed hierarchy-aware attention, explaining how to impose hierarchy aggregation priors on $C$ in three steps.

\emph{Generate neighboring attention score.} The merging tendency of adjacent word tokens is described through neighboring attention scores. Two learnable key and query matrices $W_{{Q}}^{'}$ and $W_{{K}}^{'}$ transfer any adjacent word tokens $(t_i, t_{i+1})$. The neighboring attention score $s_{i, i+1}$ is defined as their inner product:
\begin{equation}
s_{i, i+1}=\frac{(t_i W_{{Q}}^{'}) \cdot (t_{i+1} W_{{K}}^{'})}{\sigma_t}
\end{equation}
Here, $\sigma_t$ is a hyperparameter controlling the scale of the generated scores. A $\mathrm{softmax}$ function for each token $t_i$ is employed to normalize its merging tendency with two neighbors:
\begin{equation}
p_{i, i+1}, p_{i, i-1}=\operatorname{softmax}\left(s_{i, i+1}, s_{i, i-1}\right)
\end{equation}
For neighbor pairs $(t_{i},t_{i+1})$, the neighboring affinity score $\hat{a}_{i, i+1}$ is the geometric mean of $p_{i, i+1}$ and $p_{i+1, i}$: $\hat{a}_{i, i+1} = \sqrt{p_{i, i+1} \cdot p_{i+1, i}}$. In a graph perspective, it describes the strength of edge $e_{i,i+1}$ by comparing it with edges $e_{i-1, i}$ ($p_{i, i+1} \text{v.s. } p_{i,i-1}$) and $e_{i+1,i+2}$ ($p_{i+1, i} \text{v.s. } p_{i+1,i+2}$).

\emph{Enforcing Non-splittable property.} A higher neighboring affinity score indicates that two neighbor tokens are more closely bonded. To ensure that merged tokens will not be split, layer-wise affinity scores $a_{i,i+1}^l$ should increase as the network goes deeper, i.e., $a_{i,i+1}^l\geq a_{i,i+1}^{l-1}$ for all $l$. It helps gradually generate a desired hierarchy structure:
\begin{equation}
a_{i, i+1}^l=a_{i, i+1}^{l-1}+\left(1-a_{i, i+1}^{l-1}\right) \hat{a}_{i, i+1}^l
\end{equation}
%
% \emph{Modeling the tendency to merge.}
% To measure the tendency to merge, denoted as $C_{i,j}$, for \emph{any} word token pair $(t_i, t_j)$, we propagate the affinity scores of neighboring tokens between $(t_i, t_j)$. Specifically,  $C_{i,j}$ is derived through the multiplication operation as $C_{i, j}=\prod_{k=i}^{j-1} a_{k, k+1}$. Note that $C$ is a symmetric matrix, so $C_{i,j} = C_{j,i}$.

Similarly, we formulate the \emph{Hierarchy Induction for Visual Branch}, detailed in the \emph{Appendix}.

\subsection{Feature Flow Attention Mechanism}

In the following representation, we use the corresponding nodes ${h}$ from the respective branches of language and visual graph transformers as the queries ($\mathbf{Q}$), keys ($\mathbf{K}$) and values ($\mathbf{V}$).

Inspired by flow network theory, the Flow-Attention mechanism introduces a competitive mechanism for sources and an allocation mechanism for sinks, preventing the generation of trivial attention.
In a flow network framework, attention is viewed as the flow of information from sources to sinks. The results ($\mathbf{R}$) act as endpoints receiving inbound information flow, and the values ($\mathbf{V}$) serve as sources providing outgoing information flow.

\noindent\textbf{Flow Capacity Calculation.} 
For a scenario with \textit{\textbf{n}} sinks and \textit{\textbf{m}} sources, incoming flow ${I}_{i}$ for the $i$-th sink and outgoing flow ${O}_{j}$ for the $j$-th source are calculated as:
\begin{equation}
\begin{split}\label{equ:in_out_flow}
{I}_{i} &= \phi(\mathbf{Q}_{i})\sum_{j=1}^{m}\phi(\mathbf{K}_{j})^{\sf T},\\
{O}_{j} &=\phi(\mathbf{K}_{j})\sum{i=1}^{n}\phi(\mathbf{Q}_{i})^{\sf T},
\end{split}
\end{equation}
where $\phi(\cdot)$ is a non-negative function.

\noindent\textbf{Flow Conservation.}
We establish the preservation of incoming flow capacity for each sink, maintaining the default value at 1, effectively "locking in" the information forwarded to the next layer. This conservation strategy ensures that the outgoing flow capacities of sources engage in competition, with their collective sum strictly constrained to 1. Likewise, by conserving the outgoing flow capacity for each source at the default value of 1, essentially "fixing" the information acquired from the previous layer, the conservation of incoming and outgoing flow capacities is enforced by normalizing operations:
\begin{equation}
\begin{split}\label{equ:reweighting}
\frac{\phi(\mathbf{K})}{\mathbf{O}}, \frac{\phi(\mathbf{Q})}{\mathbf{I}},
\end{split}
\end{equation}
In this context, the ratio denotes element-wise division, with $\frac{\phi(\mathbf{K})}{\mathbf{O}}$ dedicated to source conservation and $\frac{\phi(\mathbf{Q})}{\mathbf{I}}$ assigned for sink conservation.

This normalization process ensures the preservation of flow capacity for each source and sink token, as evidenced by the following equations:
\begin{equation}
	\begin{split}\label{equ:conservation_proof}
	    \text{source-$j$:}\ & \frac{\phi(\mathbf{K}_{j})^{\sf T}}{{O}_{j}}\sum_{i=1}^{n}\phi(\mathbf{Q}_{i})=\frac{\sum_{i=1}^{n}\phi(\mathbf{Q}_{i})\phi(\mathbf{K}_{j})^{\sf T}}{{O}_{j}}=1 \\
	    \text{sink-$i$:}\ &
	    \frac{\phi(\mathbf{Q}_{i})^{\sf T}}{{I}_{i}}\sum_{j=1}^{m}\phi(\mathbf{K}_{j})=\frac{\sum_{j=1}^{m}\phi(\mathbf{K}_{j})\phi(\mathbf{Q}_{i})^{\sf T}}{{I}_{i}}=1\\
	\end{split}
\end{equation}

These equations replicate the same computations as~\cref{equ:in_out_flow}. The initial equation concerns the outgoing flow capacity of the $j$-th source after the normalization process $\frac{\phi(\mathbf{K})}{\mathbf{O}}$. In contrast, the second equation corresponds to the incoming flow capacity of the $i$-th sink after the normalization process $\frac{\phi(\mathbf{Q})}{\mathbf{I}}$. In both instances, the capacities are identical to the default value of 1.

The conserved incoming flow $\widehat{\mathbf{I}}$ and outgoing flow $\widehat{\mathbf{O}}$ are represented as:
\begin{equation}
\begin{split}\label{equ:conservation}
\widehat{\mathbf{I}} &= \phi(\mathbf{Q})\sum_{j=1}^{m}\frac{\phi(\mathbf{K}{j})^{\sf T}}{{O}{j}},\\
\widehat{\mathbf{O}} &= \phi(\mathbf{K})\sum_{i=1}^{n}\frac{\phi(\mathbf{Q}{i})^{\sf T}}{{I}{i}}
\end{split}
\end{equation}
\noindent\textbf{Flow-Attention Mechanism.}
We introduce the Flow-Attention mechanism, leveraging competition induced by incoming flow conservation for sinks. In $\widehat{\mathbf{O}}$, sources compete while maintaining a fixed flow capacity sum, revealing source significance. $\widehat{\mathbf{I}}$ represents sink information when source outgoing capacity is 1, reflecting aggregated information allocation to each sink. The Flow-Attention equations are:
\begin{equation}
\begin{split}\label{equ:overall}
\text{Competition:}\ &\widehat{\mathbf{V}}=\operatorname{Softmax}(\widehat{\mathbf{O}})\odot\mathbf{V}\\
\text{Aggregation:}\ &\mathbf{A}=\frac{\phi(\mathbf{Q})}{\mathbf{I}}\big(\phi(\mathbf{K})^{\sf T}\widehat{\mathbf{V}}\big)\\
\text{Allocation:}\ &\mathbf{R}=\operatorname{Sigmoid}(\widehat{\mathbf{I}})\odot \mathbf{A}.
\end{split}
\end{equation}
In the "Competition" stage, $\widehat{\mathbf{V}}$ is determined through the application of the Softmax function to $\widehat{\mathbf{O}}$, followed by element-wise multiplication with $\mathbf{V}$. The "Aggregation" step, denoted as $\mathbf{A}$, is computed using the presented equation. Lastly, the "Allocation" phase calculates $\mathbf{R}$ by employing the Sigmoid function on $\widehat{\mathbf{I}}$, which is then element-wise multiplied with $\mathbf{A}$.
\subsection{Training Loss}
To adapt the contrastive pretraining objective for video and text features in the HAtt-Flow architecture, the objective function can be expressed as follows:

\begin{align}
\mathscr{L} = -\frac{1}{M} \sum_{i}^{M} \log \frac{\exp \left(\mathbf{v}{i}^{\top} \mathbf{u}{i} / \tau\right)}{\sum_{j=1}^{M} \exp \left(\mathbf{v}{i}^{\top} \mathbf{u}{j} / \tau\right)} \notag\\ 
- \frac{1}{M} \sum_{i}^{M} \log \frac{\exp \left(\mathbf{u}{i}^{\top} \mathbf{v}{i} / \tau\right)}{\sum_{j=1}^{M} \exp \left(\mathbf{u}{i}^{\top} \mathbf{v}{j} / \tau\right)}
\end{align}

Here, $\mathbf{v}$ and $\mathbf{u}$ represent the video and text feature vectors, $\tau$ is the learnable temperature parameter, and $M$ is the total number of video-text pairs, i.e., the total number of labels.
\section{Experimental Results}\label{sec:exp}

\subsection{Experiment Settings}

\noindent \textbf{Dataset Details.} 
Our dataset adopts a division strategy from JRDB~\cite{martin2021jrdb}, where videos are segregated at the sequence level, ensuring the entirety of a video sequence is allocated to a specific split. The 54 video sequences are distributed, with 20 for training, 7 for validation, and 27 for testing. To align with the evaluation practices of analogous datasets, our evaluation is centered on keyframes sampled at one-second intervals, resulting in 1419 training samples, 404 validation samples, and 1802 test samples.

\noindent \textbf{Implementation Details.} Our framework, implemented in PyTorch, undergoes training on a machine featuring four NVIDIA Quadro RTX 6000 GPUs. During training, we adopt a batch size of 2 and leverage the Adam Optimizer, commencing the training process with an initial learning rate set at 0.0001.

\noindent \textbf{Evaluation Metrics.} We evaluate the model on two tasks: 1.) Predicate Classification (PredCls) and 2.) Video Scene Graph Generation (VSGG).
The Video Scene Graph Generation (VSGG) task aims to generate descriptive triplets for an input video. Each triplet, denoted as ($r_i, t_1, t_2, o_s, {m_s}^{(t_{1},t_{2})}, o_o, {m_o}^{(t_{1},t_{2})}$), consists of a relation $r_i$ occurring between time points $t_1$ and $t_2$, connecting a subject $o_s$ (class category) with mask tube ${m_s}^{(t_{1},t_{2})}$, and an object $o_o$ with mask tube ${m_o}^{(t_{1},t_{2})}$.
Evaluation metrics for PredCls and VSGG adhere to Scene Graph Generation (SGG) standards, utilizing Recall@K (R@K) and mean Recall@K (mR@K). Successful recall for a ground-truth triplet ($\hat{o}_s, {\hat{m}_s}^{(\hat{t}_{1}, \hat{t}_{2})}, \hat{o}_o, {\hat{m}_o}^{(\hat{t}_{1},\hat{t}_{2})}, {\hat{r}_i}^{(\hat{t}_{1},\hat{t}_{2})}$) requires accurate category labels and IOU volumes between predicted and ground-truth mask tubes above 0.5. The soft recall is recorded when these criteria are met, considering the time IOU between predicted and ground-truth intervals.
% It is important to note the subtle differences in PVSG metrics compared to VidSGG metrics for VidOR. In cases where, for example, a child repeatedly stops and goes in a video, PVSG metrics only consider the triplet once, albeit with a dispersed period. This deviation from VidOR, which accounts for several "child-1 walking on ground" triplets, helps prevent certain relations from dominating the metrics through repetition.

% \usepackage{multirow}
\begin{table}[ht!]
\centering
\caption{{Comparison with SOTA methods on \textbf{GASG dataset}.} }
\resizebox{\columnwidth}{!}{
\begin{tabular}{c|c|cc|cc|cc}
\hline
\multirow{2}{*}{\textbf{Method}} & \multirow{2}{*}{\textbf{\begin{tabular}[c]{@{}c@{}}Modality\\ (X)\end{tabular}}} & \multicolumn{2}{c|}{\textbf{R/mR@20}}              & \multicolumn{2}{c|}{\textbf{R/mR@50}}     & \multicolumn{2}{c}{\textbf{R/mR@100}} \\ \cline{3-8} 
                                 &                                                                                  & \textit{\textbf{PredCls}} & \textit{\textbf{VSGG}} & \textit{\textbf{PredCls}} & \textbf{VSGG} & \textbf{PredCls}    & \textbf{VSGG}   \\ \hline
                                 \multirow{2}{*}{IMP~\cite{xu2017scene}}             & Image                                                                            &           31.9/9.55               & 16.5/6.52              &     36.8/10.9                      & 18.2/7.05     &       38.9/11.6              & 18.6/7.23 \\
                                 & Video                                                                            & -                         & -                      & -                         & -             & -                   & -               \\
                                 \multirow{2}{*}{MOTIFS~\cite{zellers2018neural}}             & Image& 44.9/20.2& 20.0/9.10& 50.4/20.1& 21.7/9.57&   52.4/22.9    & 22.0/9.67       \\
                                 & Video & -                         & -                      & -                         & -             & -                   & -               \\
                                 \multirow{2}{*}{VCTree~\cite{tang2019learning}}             & Image&45.3/20.7 & 20.6/9.70            &    50.8/22.6& 22.1/10.2     &     52.7/23.3& 22.5/10.2       \\
                                 & Video                                                                            & -                         & -                      & -                         & -             & -                   & -               \\
                                 \multirow{2}{*}{GPSNet~\cite{lin2020gps}}             & Image&31.5/13.2 & 17.8/2.03& 39.9/16.4& 19.6/7.49     &44.7/18.3& 20.1/7.67       \\
                                 & Video                                                                            & -                         & -                      & -                         & -             & -                   & -               \\
\multirow{2}{*}{PSG~\cite{yang2022panoptic}}             & Image                                                                            &                          - & 31.4/16.2              &     -                      & 32.9/21.5     &       -              & 36.1/22.7       \\
                                 & Video                                                                            & -                         & -                      & -                         & -             & -                   & -               \\
\multirow{2}{*}{PVSG~\cite{yang2023panoptic}}            & Image                                                                            &                    -       & 38.3/18.1              &                          - & 41.7/20.8     &                   -  & 43.2/23.7       \\
                                 & Video                                                                            & -                          &         \textbf{13.6/10.2}              &                       -    &    19.2/11.1           &       -              &       26.5/14.7          \\
\multirow{2}{*}{\textbf{Ours}}            & Image                                                                            &           \textbf{57.4/35.2}               &\textbf{42.2/21.4}              &  \textbf{60.2/36.1}                       & \textbf{44.5/23.1}     &            \textbf{63.7/39.5}      & \textbf{48.1/26.9}       \\
                                  & Video&                \textbf{27.1/14.3}           & 11.2/9.1                       & \textbf{29.5/17.71}& \textbf{19.6/12.3}& \textbf{41.7/24.2}& \textbf{30.2/18.1}                \\ \hline
\end{tabular}} \label{tab:sota_ours} 
\end{table}

% \usepackage{multirow}
\begin{table}[ht!]
\centering
\caption{{Comparison with SOTA methods on \textbf{PSG dataset}.}}
\resizebox{\columnwidth}{!}{
\begin{tabular}{c|cc|cc|cc}
\hline
\multirow{2}{*}{\textbf{Methods}} & \multicolumn{2}{c|}{\textbf{R/mR@20}}              & \multicolumn{2}{c|}{\textbf{R/mR@50}}     & \multicolumn{2}{c}{\textbf{R/mR@100}} \\ \cline{2-7} 
                                  & \textit{\textbf{PredCls}} & \textit{\textbf{VSGG}} & \textit{\textbf{PredCls}} & \textbf{VSGG} & \textbf{PredCls}    & \textbf{VSGG}   \\ \hline
IMP~\cite{xu2017scene}                               & 30.5/8.97                         &           17.9/7.35            & 35.9/10.5                        &     19.5/7.88         & 38.3/11.3                   & 20.1/8.02              \\
MOTIFS~\cite{zellers2018neural}                            & 45.1/19.9                        & 20.9/9.60                      & 50.5/21.5                         & 22.5/10.1             & 52.5/22.2                   & 23.1/10.3 \\
VCTree~\cite{tang2019learning}                            & 45.9/21.4                         &      21.7/9.68                 & 51.2/23.1                        & 23.3/10.2             & 53.1/23.8                   & 23.7/10.3               \\
GPSNet~\cite{lin2020gps}                            & 38.8/17.1                         &   18.4/6.52                    & 46.6/20.2                         & 20.0/6.97             & 50.0/21.3                   & 20.6/7.2               \\
PSG~\cite{yang2022panoptic}                               & -                         & 28.2/15.4                      & -                         & 32.1/20.3             & -                   & 35.3/21.5               \\
\textbf{Ours}                     & \textbf{52.4/25.6}                & \textbf{32.9/18.4}             & \textbf{56.1/28.3}                & \textbf{35.3/21.6}    & \textbf{62.7/32.12}          & \textbf{41.34/23.1}     
\end{tabular}}\label{tab:sota_psg}
\end{table}

\subsection{Comparison with State-of-the-art} We present our comparisons with state-of-the-art methods in~\cref{tab:sota_ours} and~\cref{tab:sota_psg} for our dataset and PSG dataset. In direct comparison with the methods above, the HAtt-Flow model exhibits a notable performance advantage, establishing itself as the current state-of-the-art. This superiority is attributed to its proficiency in capturing intricate social activities among subjects across spatial and temporal dimensions. On the GASG dataset, our proposed method outperforms existing SGG methods by a significant margin on all metrics except for the R/mR@20 of the VSGG task. On the PSG dataset, it is evident that the proposed method dominated the other methods to demonstrate state-of-the-art performance.

 \subsection{Ablation Study}

 \noindent \textbf{Flow Attention Direction.} 
 We introduced a novel flow attention mechanism between the hierarchical transformers handling text and vision. To explore the impact of the flow direction between these networks, we conducted experiments as detailed in~\cref{tab:ablation1}. Our findings validate that optimal results are achieved when attention flows from the text to the vision transformer. Conversely, performance declines in the opposite direction, notably when no attention flows. This observation suggests that the cross-attention mechanism enhances the model's contextual learning capacity primarily when the flow is from text to vision because text often provides high-level semantic information and context that can guide the understanding of visual content.

\noindent \textbf{Importance of Hierarchy awareness.} 
We incorporated hierarchy awareness into the transformer framework to bolster the model's scene graph generation capabilities. Experimental results, detailed in~\cref{tab:ablation_hierarchy}, affirm that hierarchical awareness optimally enhances scene graph generation. Conversely, performance declines in the absence of this design. It is likely because when the model is aware of the hierarchy during the generation of video scene graphs, it accurately predicts all relevant nodes and their relationships (edges).

\noindent \textbf{Attributes Analysis.} 
In~\cref{tab:ablation_attributes}, we evaluate the impact of different attributes in the dataset on the model's performance in scene graph generation. Our findings affirm that including all attributes results in optimal scene graph generation. Conversely, the performance exhibits a decline when we consider individual attribute at a time. We can clearly observe the performance is proportional to the attribute distribution as shown in~\cref{fig:stats} b.) i.e., more the number of attributes better the performance. It underscores the significance of leveraging all attributes, indicating that they collectively enhance the model's capacity to grasp intricate contexts, enabling accurate scene graph generation.

\begin{table}[ht!]
\centering
\caption{{Ablation Study for \textbf{Flow Attention direction}.}}
\resizebox{\columnwidth}{!}{
\begin{tabular}{c|c|cc|cc|cc}
\hline
\multirow{2}{*}{\textbf{\begin{tabular}[c]{@{}c@{}}Flow Attention\\ Direction\end{tabular}}} & \multirow{2}{*}{\textbf{\begin{tabular}[c]{@{}c@{}}Modality\\ (X)\end{tabular}}} & \multicolumn{2}{c|}{\textbf{\textit{R/mR@20}}} & \multicolumn{2}{c|}{\textbf{\textit{R/mR@50}}} & \multicolumn{2}{c}{\textbf{\textit{R/mR@100}}} \\ \cline{3-8} 
                                                                                             &                                                                                  & \textbf{\textit{PredCls}}    & \textbf{\textit{VSGG}}   & \textbf{\textit{PredCls}}    & \textit{\textbf{VSGG}}   & \textbf{\textit{PredCls}}        & \textbf{\textit{VSGG}}        \\ \hline
\multirow{2}{*}{T $\nrightarrow$ X}                                                                     & Image&       32.1/17.4              &        18.3/9.2         &     37.15/19.2                & 21.4/10.5       &  41.6/21.8              &    23.5/14.7         \\
                                                                                             & Video & 10.2/5.3 & 4.7/2.4& 12.7/7.5 & 7.1/4.2 &   16.2/10.8& 9.8/7.3 \\
\multirow{2}{*}{X $\rightarrow$ T}                                                           & Image&    45.3/23.84                 &           20.4/10.8      &      48.1/25.6               &23.5/12.4        & 52.8/27.3               &     29.2/16.1        \\
                                                                                             & Video& 15.4/7.1 & 7.2/4.7 & 18.3/9.4 & 9.7/6.2& 24.7/12.1 & 14.1/9.4 \\
\multirow{2}{*}{T $\rightarrow$ X}                                                                                                                                     & Image& \textbf{57.4/35.2}                         & \textbf{42.2/21.4}                      & \textbf{60.2/36.1}                         & \textbf{44.5/23.1}             & \textbf{63.7/39.5}                   & \textbf{48.1/26.9}               \\
                                                                                        & Video& \textbf{27.1/14.3}           & \textbf{11.2/9.1}                       & \textbf{29.5/17.71}& \textbf{19.6/12.3}& \textbf{41.7/24.2}& \textbf{30.2/18.1} \\ \hline
\end{tabular}}\label{tab:ablation1}
\end{table}
% \usepackage{multirow}
% \begin{table}[ht!]
% \centering
% \caption{\textbf{Ablation study for attributes.} \textcolor{red}{Will be updated.}}
% \resizebox{\columnwidth}{!}{
% \begin{tabular}{c|ccc}
% \hline
% \multirow{2}{*}{\textbf{Attributes}} & \multicolumn{3}{c}{\textbf{GAVSG Metrics}}              \\ \cline{2-4} 
%                                      & \textit{R/mR@20} & \textit{R/mR@50} & \textit{R/mR@100} \\ \hline
% Interaction                          & 2.1/1/4          &                  &                   \\
% Position                             & 1.4/3.1          &                  &                   \\
% Appearance                           &                  &                  &                   \\
% Relationship                         &                  &                  &                   \\
% Situation                            &                  &                  &                   \\ \hline
% \textit{All together}                         &                  &                  &                   \\ \hline
% \end{tabular}}
% \end{table}

% \usepackage{multirow}
\begin{table}[ht!]
\centering
\caption{Ablation Study for \textbf{Hierarchy Awareness.} }
\resizebox{\columnwidth}{!}{
\begin{tabular}{c|c|cc|cc|cc}
\hline
\multirow{2}{*}{\textbf{\begin{tabular}[c]{@{}c@{}}Hierarchy\\ Awareness\end{tabular}}} & \multirow{2}{*}{\textbf{\begin{tabular}[c]{@{}c@{}}Modality\\ (X)\end{tabular}}} & \multicolumn{2}{c|}{\textit{\textbf{R/mR@20}}}              & \multicolumn{2}{c|}{\textit{\textbf{R/mR@50}}}     & \multicolumn{2}{c}{\textit{\textbf{R/mR@100}}} \\ \cline{3-8} 
                                                                                        &                                                                                  & \textit{\textbf{PredCls}} & \textit{\textbf{VSGG}} & \textit{\textbf{PredCls}} & \textit{\textbf{VSGG}} & \textit{\textbf{PredCls}}    & \textit{\textbf{VSGG}}   \\ \hline
\multirow{2}{*}{\xmark}& Image& 46.1/22.3                         & 23.7/12.5                      & 47.2/24.1                         & 24.5/13.4             & 52.6/28.3                   & 29.8/16.1               \\
                                                                                        & Video& 17.5/11.3& 10.4/8.7& 22.1/14.1& 14.1/9.61& 27.4/15.2& 25.7/16.3    
                                                                                        
                                \\
\multirow{2}{*}{\cmark}                                                  & Image& \textbf{57.4/35.2}                         & \textbf{42.2/21.4}                      & \textbf{60.2/36.1}                         & \textbf{44.5/23.1}             & \textbf{63.7/39.5}                   & \textbf{48.1/26.9}               \\
                                                                                        & Video& \textbf{27.1/14.3}           & \textbf{11.2/9.1}                       & \textbf{29.5/17.71}& \textbf{19.6/12.3}& \textbf{41.7/24.2}& \textbf{30.2/18.1} \\ \hline                        
\end{tabular}}\label{tab:ablation_hierarchy}
\end{table}

\begin{table}[ht!]
\centering
 \caption{{Ablation study for \textbf{Attributes} in our dataset.}}
 \resizebox{\columnwidth}{!}{
\begin{tabular}{c|cc|cc|cc}
\hline
\multirow{2}{*}{\textbf{Attributes}} & \multicolumn{2}{c|}{\textbf{\textit{R/mR@20}}} & \multicolumn{2}{c|}{\textbf{\textit{R/mR@50}}} & \multicolumn{2}{c}{\textbf{\textit{R/mR@100}}} \\ \cline{2-7} 
                                     & \textbf{\textit{PredCls}}    & \textbf{\textit{VSGG}}   & \textbf{\textit{PredCls}}    & \textbf{\textit{VSGG}}   & \textbf{\textit{PredCls}}        & \textbf{\textit{VSGG}}        \\ \hline
Appearance                          &    14.5/2.3          &     1.4/0.3            & 17.2/6.1                    &    2.3/0.5    &     21.1/9.1           &    7.8/2.1         \\
Relationship                             &     13.2/1.8         &    1.3/0.5             &  16.4/5.2                   &    2.1/0.4    &    20.8/8.7            &    7.5/2.4         \\
Interaction                           &      11.4/1.4               &   0.9/0.4          & 14.7/4.1                   &    1.7/0.7    &    17.7/6.4            &    6.2/1.6         \\
Position                         &      8.1/0.7               &    1.5/0.51             &  10.6/3.7                   &    1.4/0.4    &     14.2/4.7           &      4.1/0.9       \\
Situation                          &     5.7/0.2                &  0.7/0.2               &   7.2/1.7                  &    0.9/0.3    &      10.1/2.8          &   2.8/0.4          \\ \hline
All together                         &\textbf{27.1/14.3}           & \textbf{11.2/9.1}                       & \textbf{29.5/17.71}& \textbf{19.6/12.3}& \textbf{41.7/24.2}& \textbf{30.2/18.1}\\ \hline
\end{tabular}}\label{tab:ablation_attributes}
\end{table}

\begin{figure*}[ht!]
    \centering
    \includegraphics[width=1.0\textwidth]{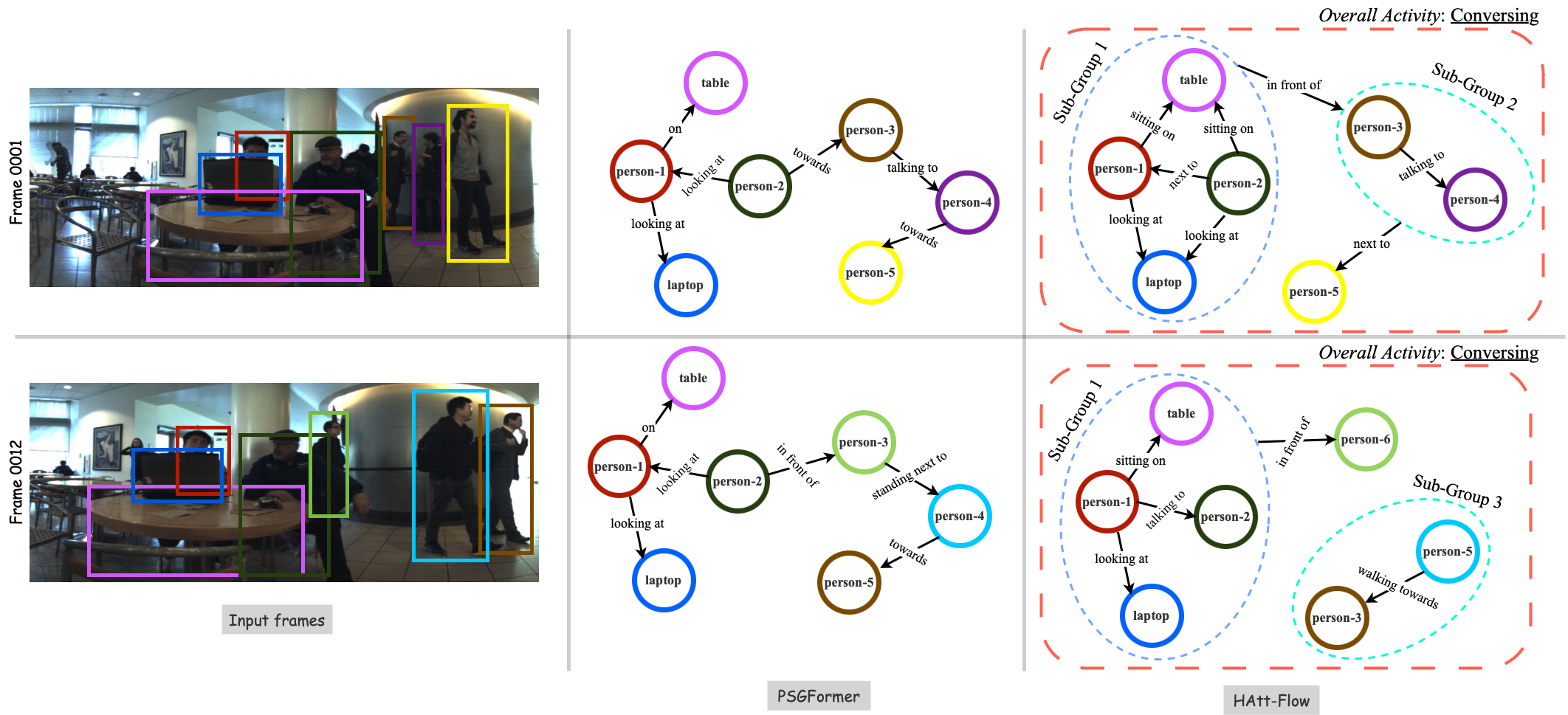}
    \caption{
    {\textbf{The visualization of the scene graphs generated by PSGFormer~\cite{yang2022panoptic} and Ours.} We can observe that ~\cite{yang2022panoptic} could only detect the subjects but not accurate groups and their interactions. In contrast, the HAtt-Flow is accurate in graph generation and overall group activity prediction. \textbf{Best viewed in color and zoomed.}}}
    \label{fig:vis}
\end{figure*}

\section{Qualitative Analysis}\label{sec:vis}
To gain deeper insights into the performance of HAtt-Flow, we employ visualization to illustrate its scene graph generation predictions using our dataset. As depicted in~\cref{fig:vis}, our model substantially improves overall scene graph generation compared to PSGFormer. The effectiveness of our hierarchy-aware attention-flow mechanism contributes significantly to this enhancement, providing our model with superior context modeling capabilities for visual representations guided by textual inputs.

\section{Conclusions}\label{sec:conclusion}
In this work, we introduced a pioneering dataset designed with nuanced attributes, specifically tailored to enhance the scene graph generation task within the context of group activities. Our contributions extend to advancing predictive video scene understanding, propelled by the introduction of the Flow-Attention and a paradigm shift in attention mechanisms through hierarchy awareness. Through rigorous experimentation, we demonstrate the efficacy of our approach, showcasing significant improvements over existing methods. 

\noindent \textbf{Limitations:} 
% The proposed research may have limitations, including increased computational demands due to the Flow-Attention mechanism. It relies heavily on the quality and quantity of training data, making it sensitive to data availability. Its effectiveness might be domain-specific and less generalizable. The model's interpretability could be reduced, and its implementation may require additional resources, limiting its use in resource-constrained applications. Scalability with complex video data and integration into existing systems could pose challenges. Performance in noisy or dynamic video environments might be limited, and latency could hinder real-time applications. Additionally, the model's applicability might be constrained in cases where obtaining accurate human annotations is difficult or expensive.
\normalsize{This work presents a novel Flow-Attention mechanism inspired by flow network theory, which introduces competition and allocation principles for attention modeling, preventing trivial attention patterns. While offering promising advancements, this approach presents some limitations. Firstly, implementing Flow-Attention entails heightened computational demands, potentially restricting its use in resource-constrained settings. Additionally, the model's performance is notably sensitive to the quality and quantity of training data, necessitating substantial data resources for optimal results. 
Balancing these computational and data requirements is essential to maximize the practicality and effectiveness of the proposed Flow-Attention mechanism in real-world applications.}
{
    \small
    \bibliographystyle{ieeenat_fullname}
    \bibliography{main}
}


\end{document}